\documentclass[sigconf]{acmart}
\AtBeginDocument{%
  }

\setcopyright{acmlicensed}
\copyrightyear{2026}
\acmYear{2026}
\setcopyright{cc}
\setcctype{by}
\acmConference[DAC '26]{63rd ACM/IEEE Design Automation Conference}{July 26--29, 2026}{Long Beach, CA, USA}
\acmBooktitle{63rd ACM/IEEE Design Automation Conference (DAC '26), July 26--29, 2026, Long Beach, CA, USA}
\acmDOI{10.1145/3770743.3807769}
\acmISBN{979-8-4007-2254-7/2026/07}

\begin{document}

\title{Special Research Session: SatReg: Regression-based Neural Architecture Search for Lightweight Satellite Image Segmentation}

\author{Edward Humes}
\email{ehumes2@jhu.edu}
\orcid{0009-0002-3945-0116}
\affiliation{%
  \institution{Johns Hopkins University}
  \city{Baltimore}
  \state{Maryland}
  \country{USA}
}

\author{Tinoosh Mohsenin}
\email{Tinoosh@jhu.edu}
\orcid{0000-0001-5551-2124}
\affiliation{%
  \institution{Johns Hopkins University}
  \city{Baltimore}
  \state{Maryland}
  \country{USA}
}


\begin{abstract}
As Earth-observation workloads move toward onboard and edge processing, remote-sensing segmentation models must operate under tight latency and energy constraints. We present SatReg, a regression-based hardware-aware tuning framework for lightweight remote-sensing segmentation on edge platforms. Using CM-UNet as the teacher architecture, we reduce the search space to two dominant width-related variables, profile a small set of student models on an NVIDIA Jetson Orin Nano, and fit low-order surrogate models for mIoU, latency, and power. Knowledge distillation is used to efficiently train the sampled students. The learned surrogates enable fast selection of near-optimal architecture settings for deployment targets without exhaustive search. Results show that the selected variables affect task accuracy and hardware cost differently, making reduced-space regression a practical strategy for adapting hybrid CNN-Mamba segmentation models to future space-edge systems.
\end{abstract}



\begin{teaserfigure}
   \centering
   \includegraphics[width=1.0\textwidth]{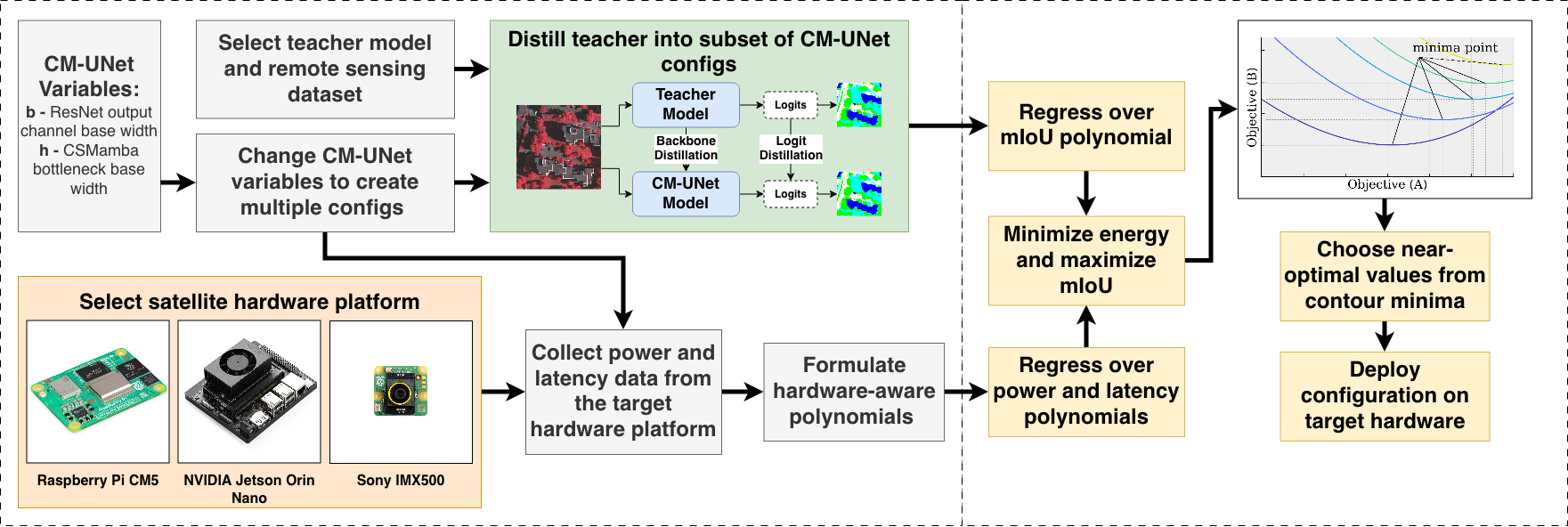}
   \caption{A high-level overview of SatReg. We vary two CM-UNet architecture parameters, profile latency/power on Jetson Orin Nano, train a subset of student models with distillation, fit surrogate regressors for mIoU and hardware cost, and optimize for near-optimal deployment points.}
   \label{fig:proposed_architecture}
\end{teaserfigure}



\maketitle

\section{Introduction}

Space-edge computing is making onboard perception increasingly important for small satellites~\cite{bouzoukis2025overview, wekerle2017status, cheng2025nanosatellite}, where power, memory, latency, and downlink limits constrain model deployment~\cite{giuffrida2021varphi, guerrisi2023artificial} on the wide array of potential hardware platforms~\cite{bui2024edge, yan2022automatic, cratere2025towards, rad2023preliminary, slater2020total}. For remote sensing, semantic segmentation is especially demanding given that it requires both fine-grained local detail and broad contextual reasoning~\cite{li2021multiattention}. Recent hybrid models\cite{vaswani2017attention, dosovitskiy2020image, xu2021efficient, wang2022novel, wang2022unetformer, gu2024mamba, dao2024transformers, zhu2024samba, ma2024rs} such as CM-UNet~\cite{liu2024cm} address this by combining convolutions with Mamba and transformer-based mechanisms, but their accuracy gains come with deployment costs that are difficult to balance on resource-constrained hardware~\cite{walczak2025bitmedvit, uttej-tecs-magrip, uttej-vlsi-2024, mozhgan-tecs-metareasoning-llm, romina-biocas-2025, uttej-biocas-2025}.

In this work, we introduce SatReg, a reduced-space regression-based tuning framework for edge-oriented deployment of CM-UNet. Rather than searching over all architectural choices, the SatReg approach (as shown in Figure \ref{fig:proposed_architecture}) searches a compact two-variable space defined by encoder width $b$ and decoder bottleneck width $h$, trains a small set of student configurations with distillation, profiles them on a Jetson Orin Nano for mIoU, latency, and power, and fits surrogate models to identify a near-optimal deployment point without brute-force exploration. The primary contributions of this work are as follows:

\begin{itemize}
\item We formulate a compact two-variable design space for CM-UNet that enables efficient exploration of accuracy, latency, and energy tradeoffs for edge-oriented remote-sensing segmentation.
\item We show that a small number of sampled configurations is sufficient to fit surrogate models that capture mIoU, latency, and power trends over this reduced space.
\item We demonstrate how the resulting surrogate-guided analysis can be used to select a near-optimal distilled CM-UNet configuration for embedded hardware relevant to future onboard deployment.
\end{itemize}
\section{Related Work}

Hardware aware model optimization for embedded AI has often relied on NAS and hardware and software co design, but these methods are expensive because many candidate models must be trained and profiled under deployment constraints. Regression based methods such as QS NAS~\cite{hosseini2021qs}, Reg Tune~\cite{mazumder2024reg}, Reg TuneV2~\cite{mazumder2023reg}, and ViT Reg~\cite{arnab-design-and-test-2024} reduce this cost by learning surrogate models for accuracy and hardware efficiency. In remote sensing, prior work has explored both NAS for semantic segmentation~\cite{wang2022dnas, broni2022evolutionary} and knowledge distillation across related tasks~\cite{le2024leveraging, chen2018training, wang2024knowledge}. SatReg brings these ideas together by applying regression guided hardware aware tuning to a hybrid CNN Mamba remote sensing segmentation model, while also using teacher-student distillation to preserve performance when tuning width scaled CM-UNet variants that cannot directly reuse pretrained weights.
\section{Method}

\begin{figure}[h!]
   \centering
   \includegraphics[width=0.8\linewidth]{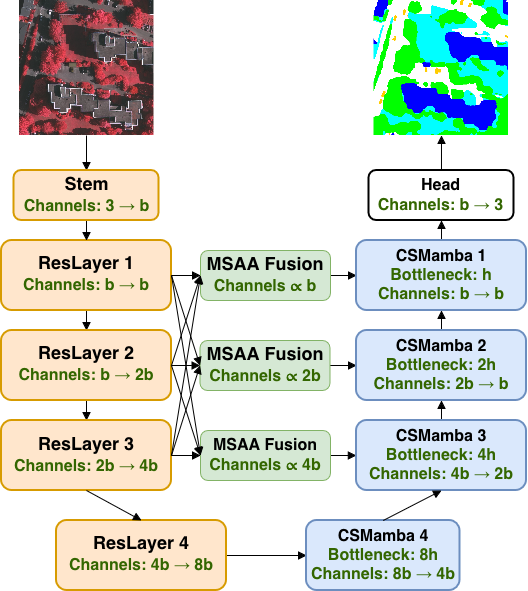}
   \caption{A high-level overview of the selected CM-UNet remote sensing image segmentation architecture. The residual blocks are used as the encoder portion of the network architecture, and the CSMamba blocks are used as the decoder, with the feature fusion and MSAA blocks sitting in between the encoder and decoder.}
   \label{fig:cmunet_architecture}
\end{figure}


SatReg targets CM-UNet, a hybrid CNN-Mamba model for remote-sensing segmentation with a ResNet-style encoder, multi-scale feature fusion, and a CSMamba decoder derived from VMamba~\cite{liu2024vmamba}, in which VSS-style selective-scan blocks are adapted to the segmentation setting. Rather than searching over layer counts or modifying the overall topology, we define a reduced design space using two architectural variables. 

The first variable, $b$, is primarily the base encoder width. It scales the channel dimensions of the four encoder stages, scales the widths of the intermediate MSAA and fusion modules that connect encoder features to the decoder, and also the decoder outer channel widths. Thus, changing $b$ scales the model's macro-level feature capacity and the cost of the backbone, decoder, and bridge modules. 

The second variable, $h$, is the base decoder bottleneck width within the CSMamba blocks. We use $h$ to define this latent bottleneck, with the per-stage bottleneck scaled proportionally across decoder stages. This decouples the expensive SS2D computation from the larger decoder channel widths, and as we found, this bottleneck width provides a direct handle on latency and power without much of an impact on mIoU.

Under this parameterization, each CM-UNet instance is indexed by a pair $(b,h)$. This preserves the original hybrid CNN-Mamba topology while exposing a compact and continuous design space for hardware-aware tuning. In contrast to encoder/decoder depth changes, which provide only coarse step-wise complexity reductions, varying $(b,h)$ enables finer control over both segmentation quality and deployment cost.

\subsection{Regression-Based Surrogate Modeling}

We profiled the various CM-UNet architectural components individually on our target hardware platform with varying values of $b$ and $h$, in order to determine latency and power trends. Predictably, we found that the power and latency trends for the ResNet~\cite{he2016deep} backbone, MSAA, and other non-decoder parts of the model are primarily controlled by $b$, whereas the decoder cost depends jointly on $b$ and $h$ because the CSMamba selective-scan blocks operate in an $h$-scaled latent space while remaining coupled to the surrounding outer encoder and decoder widths. While we found that the CSMamba blocks generally followed a third-degree polynomial trend while measuring them in isolation, when we built a third-degree polynomial for regression over our sampled models, the polynomial would overfit the measured data. We therefore utilized quadratic polynomials for latency and power:
\begin{equation}
\hat{L}(b,h)=l_0+l_1b+l_2h+l_3b^2+l_4bh+l_5h^2
\end{equation}
\begin{equation}
\hat{P}(b,h)=p_0+p_1b+p_2h+p_3b^2+p_4bh+p_5h^2.
\end{equation}

To model segmentation quality, we use the rational-form regression adopted for approximating mIoU in RegTune~\cite{mazumder2024reg}:
\begin{equation}
\hat{m}(b,h)=\frac{a_3+a_4b+a_5h+a_6bh}{a_0+a_1b+a_2h+bh}.
\end{equation}

Using these surrogates, we optimize over the continuous $(b,h)$ space with a normalized objective that favors low latency and power and high mIoU:
\begin{equation}
\begin{split}
\min_{b,h} \quad & w_L\frac{\hat{L}(b,h)-L_{\min}}{L_{\max}-L_{\min}} + w_P\frac{\hat{P}(b,h)-P_{\min}}{P_{\max}-P_{\min}} \\
& - w_m\frac{\hat{m}(b,h)-m_{\min}}{m_{\max}-m_{\min}}
\end{split}
\end{equation}
The selected continuous solution is then mapped back to the nearest valid discrete CM-UNet configuration for training and deployment.

\subsection{Distilling Student Architectures}

The baseline CM-UNet uses a ResNet-18 backbone initialized from large-scale semi-weakly supervised pretraining~\cite{yalniz2019billion}. Because changing $b$ alters the backbone channel dimensions, these weights cannot be transferred directly to width-scaled students. Instead of pretraining new variants from scratch, we therefore train each student using Channel-Wise Knowledge Distillation (CWD)~\cite{shu2021channel} on backbone features together with the segmentation loss, which helps to recover a portion of lost segmentation performance in the smaller models.
\section{Experimental Setup}

We evaluate SatReg on two remote-sensing segmentation benchmarks: LoveDA~\cite{wang2021loveda}, using the official evaluation protocol, and ISPRS Vaihingen~\cite{cramer2010dgpf, isprs_vaihingen_2d_semantic_labeling}, using the experimental settings of \cite{wang2022unetformer}. All training, inference profiling, and deployment measurements are carried out on an NVIDIA Jetson Orin Nano Super configured in MAXN SUPER mode and implemented in PyTorch. Power measurements are obtained directly from the Jetson's built-in CPU-GPU power sensor.

For each sampled architecture configuration $(b,h)$, we train the corresponding student model and measure its segmentation accuracy and hardware performance on-device. Specifically, we record mIoU, per-image inference latency, and average power consumption, and then compute energy per image from the measured latency and power. To reduce measurement noise, latency and power are each averaged over 100 inference runs, while mIoU is averaged over two independent training seeds.

During training, all models use input crops of size $512\times512$. During deployment and hardware evaluation, however, we use $1024\times1024$ images in order to better reflect the target inference setting. Because pretrained weights are not available for CM-UNet, we first train the baseline CM-UNet ourselves, report its performance as our reference model, and then use it as the teacher network for SatReg distillation. Unless otherwise noted, all training and deployment experiments are performed in FP32.
\section{Near-Optimal Configuration}

\begin{figure}
   \centering
   \includegraphics[width=0.9\linewidth]{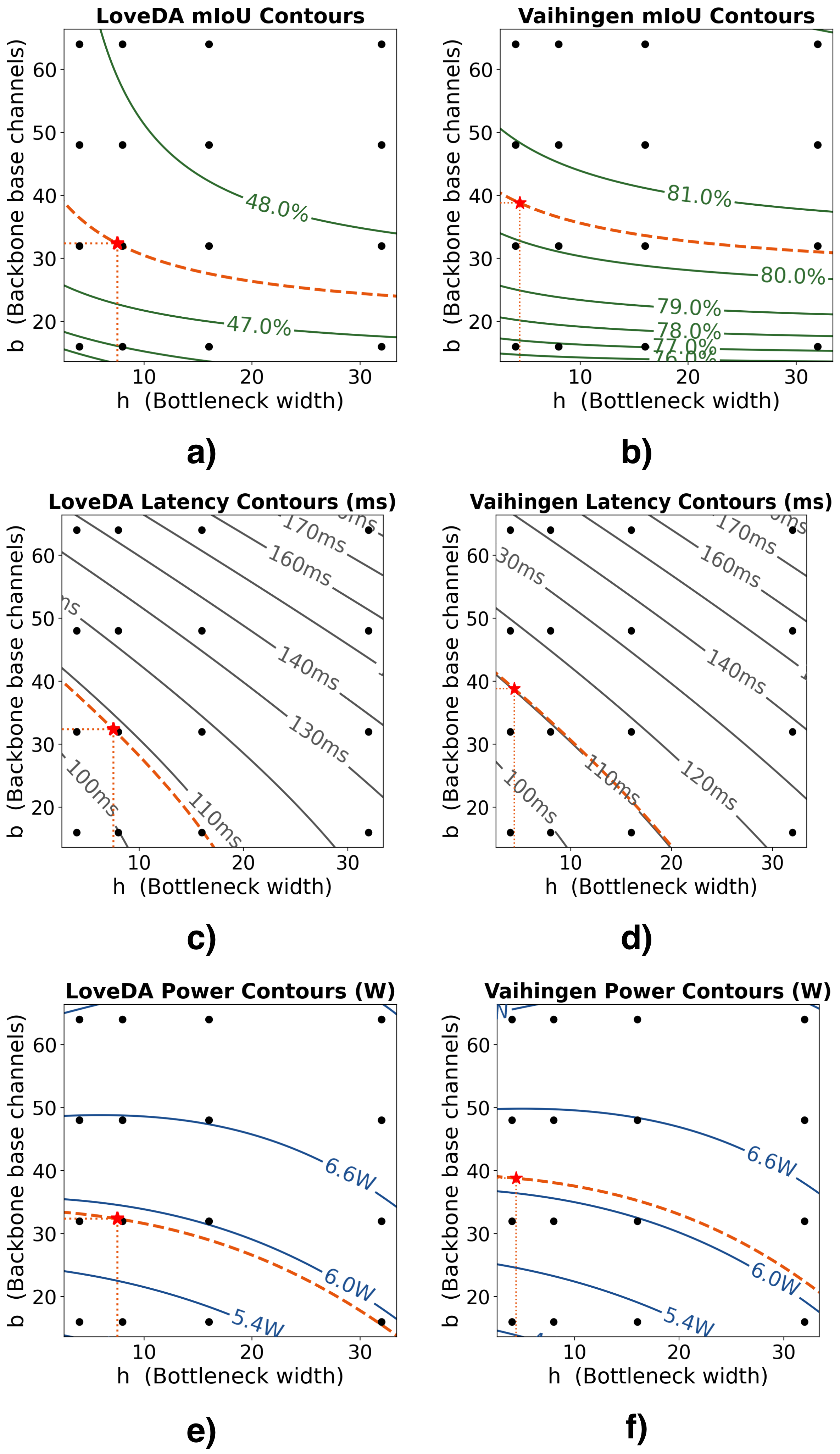}
   \caption{Contour plots of the SatReg surrogate surfaces over the reduced CM-UNet design space for LoveDA and Vaihingen. The top row shows predicted mIoU, the middle row predicted latency, and the bottom row predicted power as functions of encoder base width $b$ and decoder bottleneck width $h$. Black dots indicate sampled configurations used for regression, and orange markers/contour lines indicate the selected near-optimal configurations.}
   \label{fig:contour_plots}
\end{figure}

\begin{table}
\resizebox{\linewidth}{!}{
\begin{tabular}{|cccc|ccc|}
\hline
\multicolumn{2}{|c|}{}                                                                                                              & \multicolumn{1}{c|}{\textbf{Predicted}} & \textbf{Deployed} & \multicolumn{1}{c|}{}                 & \multicolumn{1}{c|}{\textbf{Predicted}} & \textbf{Deployed} \\ \hline
\multicolumn{1}{|c|}{\textbf{Parameters}}                                                   & \multicolumn{1}{c|}{\textbf{CM-UNet}~\cite{liu2024cm}} & \multicolumn{2}{c|}{\textbf{SatReg (Ours)}}                      & \multicolumn{1}{c|}{\textbf{CM-UNet}} & \multicolumn{2}{c|}{\textbf{SatReg (Ours)}}                      \\ \hline
\multicolumn{1}{|c|}{\textbf{Dataset}}                                                      & \multicolumn{3}{c|}{LoveDA}                                                                         & \multicolumn{3}{c|}{Vaihingen}                                                                      \\ \hline
\multicolumn{1}{|c|}{\textbf{b}}                                                            & \multicolumn{1}{c|}{64}               & \multicolumn{2}{c|}{\textbf{32}}                            & \multicolumn{1}{c|}{64}               & \multicolumn{2}{c|}{\textbf{40}}                            \\ \hline
\multicolumn{1}{|c|}{\textbf{h}}                                                            & \multicolumn{1}{c|}{32}               & \multicolumn{2}{c|}{\textbf{8}}                             & \multicolumn{1}{c|}{32}               & \multicolumn{2}{c|}{\textbf{4}}                             \\ \hline
\multicolumn{1}{|c|}{\textbf{\begin{tabular}[c]{@{}c@{}}Test\\ mIoU (\%)\end{tabular}}}     & \multicolumn{1}{c|}{50.21}            & \multicolumn{1}{c|}{47.54}              & \textbf{47.45}    & \multicolumn{1}{c|}{84.64}            & \multicolumn{1}{c|}{80.48}              & \textbf{80.64}    \\ \hline
\multicolumn{1}{|c|}{\textbf{Latency (ms)}}                                                 & \multicolumn{1}{c|}{178.63}           & \multicolumn{1}{c|}{108.20}             & \textbf{109.72}   & \multicolumn{1}{c|}{178.27}           & \multicolumn{1}{c|}{110.22}             & \textbf{119.69}   \\ \hline
\multicolumn{1}{|c|}{\textbf{Power (W)}}                                                    & \multicolumn{1}{c|}{7.21}             & \multicolumn{1}{c|}{5.90}               & \textbf{5.75}     & \multicolumn{1}{c|}{7.19}             & \multicolumn{1}{c|}{6.11}               & \textbf{6.07}     \\ \hline
\multicolumn{1}{|c|}{\textbf{\begin{tabular}[c]{@{}c@{}}Energy \\ (mJ/image)\end{tabular}}} & \multicolumn{1}{c|}{1287.92}          & \multicolumn{1}{c|}{638.38}             & \textbf{630.67}   & \multicolumn{1}{c|}{1281.76}          & \multicolumn{1}{c|}{673.73}             & \textbf{726.63}   \\ \hline
\multicolumn{1}{|c|}{\textbf{Frames/Second}}                                                & \multicolumn{1}{c|}{5.60}             & \multicolumn{1}{c|}{9.24}               & \textbf{9.11}     & \multicolumn{1}{c|}{5.60}             & \multicolumn{1}{c|}{9.09}               & \textbf{8.35}     \\ \hline
\multicolumn{1}{|c|}{\textbf{FPS/Watt}}                                                     & \multicolumn{1}{c|}{0.78}             & \multicolumn{1}{c|}{1.57}               & \textbf{1.58}     & \multicolumn{1}{c|}{0.78}             & \multicolumn{1}{c|}{1.49}               & \textbf{1.37}     \\ \hline
\multicolumn{1}{|c|}{\textbf{\begin{tabular}[c]{@{}c@{}}Model Size \\ (MB)\end{tabular}}}   & \multicolumn{1}{c|}{51.56}            & \multicolumn{2}{c|}{\textbf{12.36}}                         & \multicolumn{1}{c|}{51.56}            & \multicolumn{2}{c|}{\textbf{18.86}}                         \\ \hline
\end{tabular}
}
\vspace{1ex}
\caption{Comparison of baseline CM-UNet and our SatReg-selected near-optimal configurations on LoveDA and Vaihingen; Image size used is $1024\times1024$.}
\label{tab:near_optimal_configs}
\vspace{-4ex}
\end{table}

SatReg fits its surrogates using 16 sampled configurations per dataset over $b \in {16,32,48,64}$ and $h \in {4,8,16,32}$, achieving strong predictive agreement on LoveDA and Vaihingen, with most $R^2$ values around 97\% or higher with the exception of 89\% for the LoveDA mIoU model. It then searches the continuous (b,h) space for configurations that jointly favor high mIoU and low latency/energy, and maps the selected solution to the nearest discrete CM-UNet design. Table~\ref{tab:near_optimal_configs} summarizes the resulting points.

\begin{figure}
   \centering
   \includegraphics[width=1.0\linewidth]{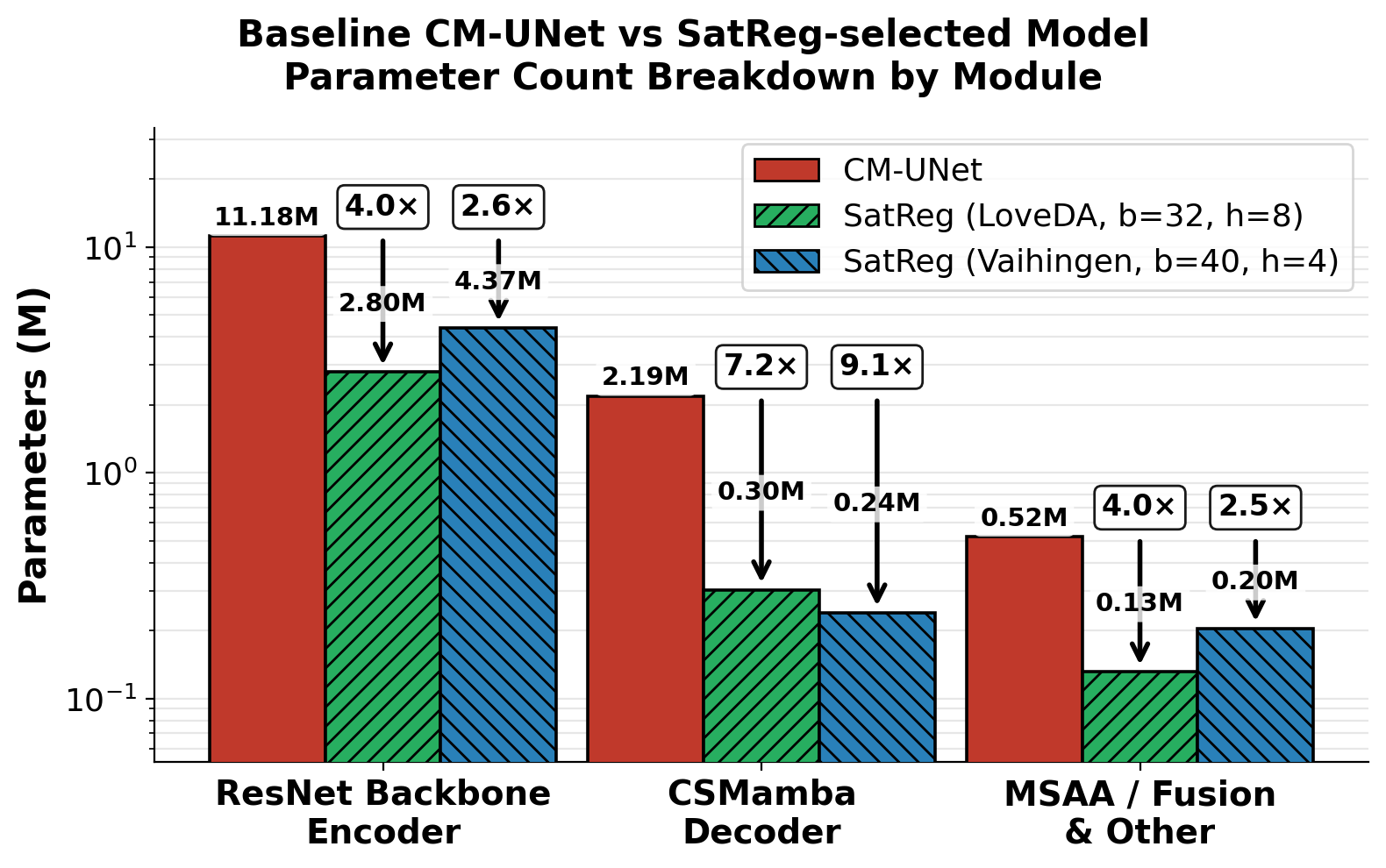}
   \caption{The parameter counts of the various modules of the baseline CM-UNet model, compared to the module parameter counts for the Vaihingen and LoveDA SatReg-selected models.}
   \label{fig:module_param_counts}
\end{figure}

\begin{figure}
   \centering
   \includegraphics[width=1.0\linewidth]{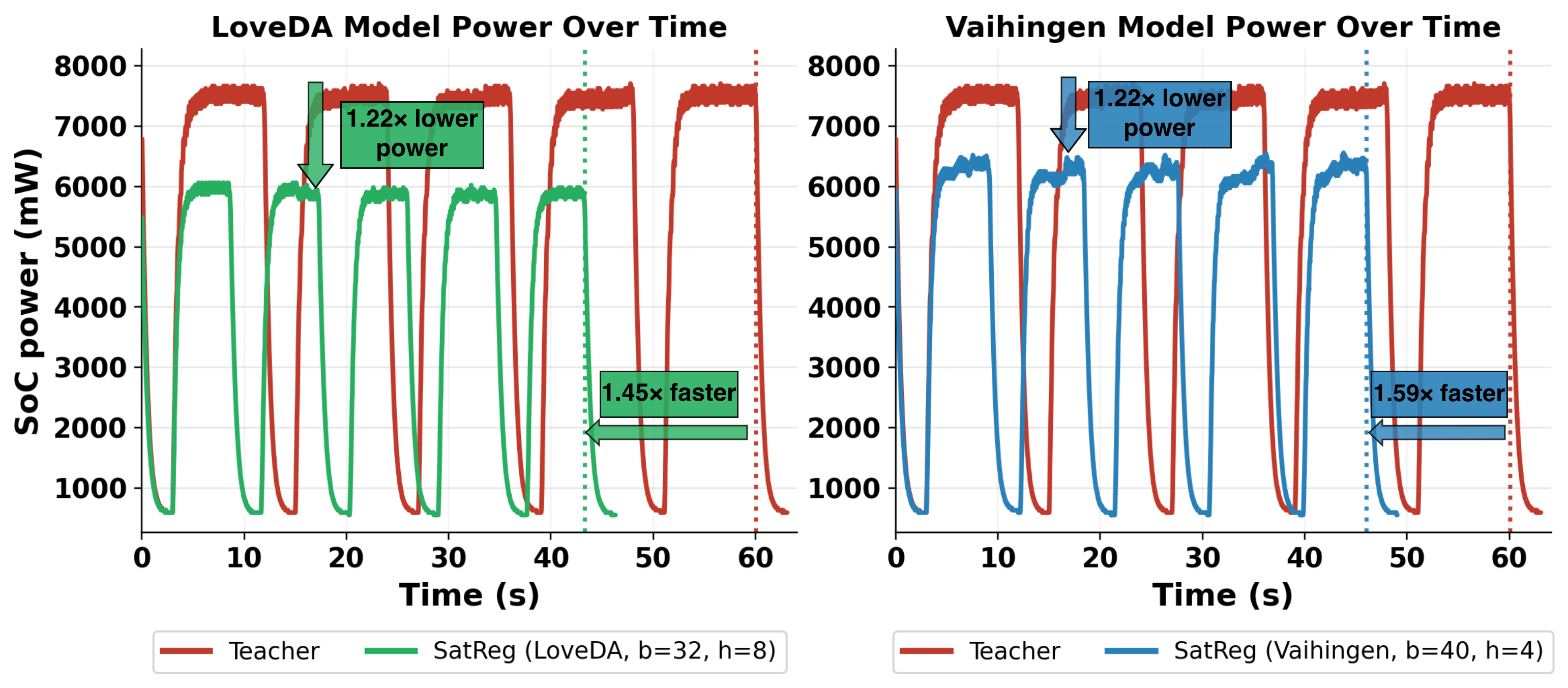}
   \caption{Jetson Orin Nano Power Versus Time for the baseline CM-UNet model and our SatReg-selected LoveDA and Vaihingen models when processing 5 batches of 50 inferences.}
   \label{fig:power_over_time}
\end{figure}

As shown in Table \ref{tab:near_optimal_configs}, SatReg reduces latency by 32.9–38.6\% and energy by 43.3–51.0\% relative to baseline CM-UNet, while retaining most of the segmentation accuracy. Throughput increases from 5.60 FPS to 9.11 FPS on LoveDA and 8.35 FPS on Vaihingen. Figure \ref{fig:power_over_time} shows corresponding reductions in execution time and power during repeated inference on Jetson Orin Nano when processing 5 batches of 50 iterations for the baseline CM-UNet model, and our two SatReg-selected models. Finally, as depicted by Figure \ref{fig:module_param_counts}, the parameter counts of the various CM-UNet components are decreased within our SatReg-selected configurations.

\begin{figure}
   \centering
   \includegraphics[width=1.0\linewidth]{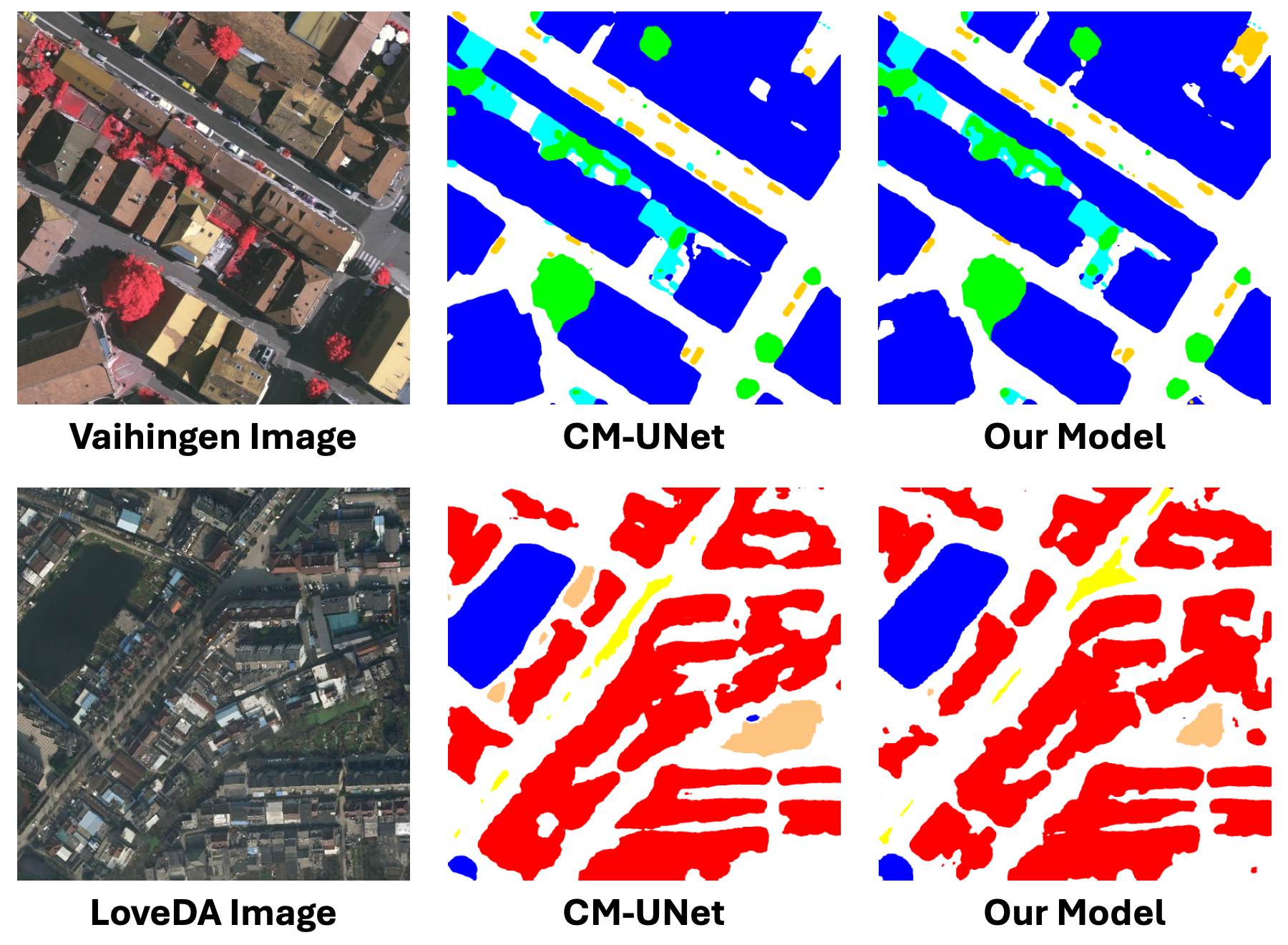}
   \caption{Qualitative comparison of segmentations by CM-UNet vs. our SatReg model configurations on LoveDA and Vaihingen.}
   \label{fig:segmentations}
\end{figure}

To contextualize these results against hybrid Mamba-based remote sensing segmentation models, we also considered RS3Mamba \cite{ma2024rs} (82.78\% mIoU on Vaihingen, 50.62\% mIoU on LoveDA) and UNetMamba~\cite{zhu2024unetmamba}(83.47\% mIoU on Vaihingen, 53.35\% mIoU on LoveDA). We note that the reported Vaihingen results for these prior works were obtained under their respective training protocols, which differ from our Vaihingen split. UNetMamba could not be executed on the Jetson Orin Nano in our PyTorch inference setting due to CUDA memory allocator limitations, while RS3Mamba was only runnable after reducing the input image resolution to 512$\times$512. At that reduced resolution, RS3Mamba drew approximately 7.95 W and required about 250 ms per image on both Vaihingen and LoveDA. 

The contour plots in Figure \ref{fig:contour_plots} indicates that mIoU varies more strongly with encoder width $b$, whereas latency and power are more sensitive to decoder bottleneck width $h$. Consequently, the selected configurations retain more encoder capacity while shrinking the CSMamba bottleneck, producing substantial deployment gains with moderate mIoU loss. Figure \ref{fig:segmentations} further shows that the selected models preserve the main scene structure and large object regions of the baseline predictions.
\section{Conclusion}


SatReg introduces a reduced-space regression framework for adapting CM-UNet to embedded space-edge platforms. By optimizing over encoder width and decoder bottleneck width, SatReg identifies near-optimal deployment points on Jetson Orin Nano without exhaustive search, reducing latency by 32.9–38.6\% and energy by 43.3–51.0\% relative to baseline CM-UNet while retaining most of the segmentation accuracy. These results demonstrate regression-guided reduced-space tuning is a practical strategy for adapting modern remote-sensing segmentation models to future space edge AI systems.
\begin{acks}
The Vaihingen data set was provided by the German Society for Photogrammetry, Remote Sensing and Geoinformation (DGPF)~\cite{cramer2010dgpf}: http://www.ifp.uni-stuttgart.de/dgpf/DKEP-Allg.html. This work was carried out at the Advanced Research Computing at Hopkins (ARCH) core facility using the DSAI cluster.
\end{acks}

\bibliographystyle{ACM-Reference-Format}
\bibliography{eehpc, refs}

\end{document}